# EXPLORING CAUSALWORLD: ENHANCING ROBOTIC MANIPULATION VIA KNOWLEDGE TRANSFER AND CURRICULUM LEARNING

**Xinrui Wang**
Dept. of Aerospace & Mechanical Engineering
University of Southern California
Los Angeles, USA
xinruiw@usc.edu

**Yan Jin***
Dept. of Aerospace & Mechanical Engineering
University of Southern California
Los Angeles, USA
yjin@usc.edu
(*corresponding author)

**ABSTRACT**

*This study explores a learning-based tri-finger robotic arm manipulating task, which requires complex movements and coordination among the fingers. By employing reinforcement learning, we train an agent to acquire the necessary skills for proficient manipulation. To enhance the efficiency and effectiveness of the learning process, two knowledge transfer strategies, fine-tuning and curriculum learning, were utilized within the soft actor-critic architecture. Fine-tuning allows the agent to leverage pre-trained knowledge and adapt it to new tasks. Several variations like model transfer, policy transfer, and across-task transfer were implemented and evaluated. To eliminate the need for pretraining, curriculum learning decomposes the advanced task into simpler, progressive stages, mirroring how humans learn. The number of learning stages, the context of the sub-tasks, and the transition timing were found to be the critical design parameters. The key factors of two learning strategies and corresponding effects were explored in context-aware and context-unaware scenarios, enabling us to identify the scenarios where the methods demonstrate optimal performance, derive conclusive insights, and contribute to a broader range of learning-based engineering applications.*

**Keywords**: Artificial intelligence, transfer learning, curriculum learning, reinforcement learning, robotic manipulation

## 1. INTRODUCTION

Robotic manipulation, a fundamental aspect of robotics research, involves various tasks where robots interact with objects to achieve specific goals, such as grasping, pushing, and stacking them. Although many studies on robotic arm manipulation focus on using grippers [4, 8, 9], which rely on actuators for basic open and close movements, our research explores the transfer of knowledge in tasks that require complex control and intricate movements. To accomplish this, we have chosen a tri-finger robotic arm manipulation task from CausalWorld [10] to conduct case studies. This setup features a three-fingered arm, with each finger having a joint and an end-effector interacting with blocks. Such a configuration demands synchronized movements for grasping and lifting objects, leading to a wide range of possible finger movements and interactions, thereby significantly increasing the task's complexity. Additionally, CausalWorld allows us to modify the characteristics of the arms and blocks, such as size, shape, weight, friction, etc. This flexibility provides us an opportunity to explore the effects of such interventions on our chosen approach, as well as to assess the generalization and robustness.

Learning-based methods have been leveraged and demonstrated significant advancements in the robotic manipulation field [37, 38]. Rather than relying solely on bootstrap learning, knowledge transfer techniques enable agents to tackle target tasks with greater efficiency and effectiveness by leveraging previously acquired knowledge. In general, knowledge can be classified into explicit and implicit categories. The explicit knowledge can be conveyed explicitly through structured information sources, such as books, manuals, procedures, or expert systems [7, 18, 19, 25]. Implicit knowledge is typically shared through task illustration or execution, which can be transferred between different tasks [20]. Our previous study explored implicit knowledge transfer within the ship collision avoidance domain. We integrated transfer learning into the reinforcement learning domain and proposed the transfer reinforcement learning methodology [21, 22]. By capturing and utilizing both image and dynamic features, the agent demonstrated adaptability to complex scenarios after being pre-trained in simpler tasks. To validate the transfer reinforcement learning approach further, we have expanded our investigation from 2D to 3D environments in this study. Moreover, our previous work employed the relatively straightforward Deep Q-Network (DQN) model [40], focusing on the whole model or layer-level transfer. In pursuit of a deeper understanding of knowledge transfer, this study utilized an actor-critic based model [35], distinguishing between the policy and value functions. The model's complex structure provides opportunities to explore knowledge transfer across its different components.

In this study, the fine-tuning approach was employed as a method for knowledge transfer, utilizing the actor-critic model to incorporate various learning strategies. Typically, the whole model is pre-trained in the source task and transferred to the target task. When the source and target tasks have misaligned reward structures or environmental dynamics. By transferring only the policy, there's a reduced risk of importing irrelevant value estimations to the target task learning. Besides the knowledge transfer across the model structure, another variation, knowledge transfer across different tasks, was explored. As humans, we are often able to easily learn a complicated task based on a related but simpler task. For example, we can master pushing blocks to arbitrary goals quickly as an add-on skill if we already know how to grasp a block. Inspired by this observation, we sought to determine whether robot learning could follow a similar logic.

To eliminate the need for additional efforts for pre-training the agent, curriculum learning was employed as another option to facilitate learning in a single iteration [39]. Curriculum learning is analogous to educational curricula in human learning, where foundational skills are developed through solving basic tasks before progressing to more advanced ones [31]. It is an extension of transfer learning, as the knowledge gained in simpler tasks is transferred to solve more complex tasks. The primary motivation for applying curriculum learning is that, by breaking down the learning process into a series of progressively challenging tasks, the agent can not only speed up the learning process but also lead to more robust and generalizable results. The curriculum design, like what sub-tasks can be involved and how to set the schedule for the subtasks, is one of the main challenges, as it directly affects the learning process and results.

Upon these, we raised the following research questions for our investigation of learning strategies for robotic manipulation:

*1. How do variations in fine-tuning, such as the whole-model or policy-only transfer and the within-task or across-task transfer, impact learning performance?*

*2. What are the critical design parameters for curriculum design, and how do these parameters influence the curriculum learning process?*

*3. Given the distinct advantages of each learning approach, how can we determine the most suitable choice for different scenarios?*

To address these questions, we designed a series of case studies to investigate the learning methods we proposed. The key factors of the learning approaches and the corresponding effect were explored, enabling us to identify the scenarios where the methods demonstrate optimal performance, thereby deriving conclusive insights. Furthermore, our findings are intended to contribute to a broader range of learning-based engineering applications.

The remaining sections of this paper are structured as follows. Section 2 provides the related work to the selected benchmark and learning methodology. Section 3 outlines the methods employed to conduct the case studies, which are detailed in Section 4. The results and discussions are covered in Section 5. The conclusion and future work are presented subsequently.

## 2. RELATED WORK

### 2.1 Reinforcement Learning in Robotic Manipulation

Recognized as a highly complex task, robotic manipulation requires a sophisticated integration of techniques from multiple disciplines like control systems, motion planning, and trajectory tracking, empowering robots to effectively interact with and manipulate objects within their surroundings [1].

Reinforcement learning, as an important method in the robotic manipulation domain, enables the robotic to learn the optimal policy from trial and error through interacting with the surrounding environment. Zhao et al. combined the control method with actor-critic and developed an RL-based controller to maintain safety and stability [23]. Wang et al. proposed an end-to-end RL framework for robotic manipulation tasks such as grasping and pushing., utilizing the state representation learned from camera images through self-supervised autoencoder [4]. Gu et al. demonstrated that an off-policy deep Q-functions-based algorithm can be applied to various complex 3D manipulation tasks in simulation and a complex door-opening skill on real robots efficiently without any prior demonstrations [5]. Liu et al. provided a review of deep reinforcement learning applications in robotic manipulation, tackling the sample efficiency and generalization challenges [6]. They also highlighted the remaining challenge of developing robust and versatile manipulation skills, pointing to the need for the integration of other machine learning to generate new algorithm solutions. Building upon this foundation, our research is focusing on improving the learning efficiency and robustness of highly complex robotic manipulation tasks. By integrating transfer learning and curriculum learning algorithms, we aim to address the gaps identified by Liu et al. and discuss the inherent knowledge and capabilities in various situations.

### 2.2 CausalWorld

CausalWorld is a causal reinforcement learning benchmark developed on the PyBullet physics engine [11], designed to facilitate research in causal structure and transfer learning within robotic manipulation tasks. This platform has become foundational for a broad spectrum of research aiming at diverse objectives. In the realm of reinforcement learning, CausalWorld plays an important role in its well-designed state and action space. Its experimental environment has been leveraged to create novel reinforcement learning models [13]. Moreover, the platform's reward function is intricately designed to address a variety of manipulation tasks, enhancing its utility and flexibility for research. Allshire et al. follow the position-orientation-based success criterion and dense reward function for object reaching of CausalWorld to conduct their reward function design for various manipulation tasks like grasping and fine corrections [14].

For studies in causal reasoning, CausalWorld also served as a crucial benchmark due to its design centered around causal tasks and the extensive space for interventions it offers. Its carefully designed tasks and the capability for random interventions have also inspired the design of other study cases in causal reasoning [12]. Meanwhile, to our knowledge, no existing research has utilized CausalWorld with the explicit aim of exploring transfer learning and curriculum learning as one of its primary objectives, particularly in examining the effects of interventions and learning curriculums as we have. The two studies most closely related to our work are Causal Curiosity [15] and Causal Counterfactuals [16]. Causal Curiosity introduced a reward mechanism that encouraged agents to learn action sequences to deduce the binary quantized representation of factors in dynamic environments, with interventions applied to manipulate these factors. The learned representation could then be concatenated to the state and transferred to downstream tasks. Inspired by the intervention utilization and causal representation concatenation, Causal Counterfactuals learned the causal representations by modifying CoPhy [17], utilizing these representations to enhance the efficiency of learning in manipulation tasks. The primary aim of these two research is to efficiently uncover causal factors to facilitate policy training, whereas our research focuses on examining how these factors affect knowledge transfer and the learning process. We also employ curriculum learning to gradually familiarize the agent with the environment in a single iteration, eliminating the need for additional efforts to identify causal representations.

### 2.3 Knowledge Transfer

Knowledge transfer aims to leverage knowledge from source domains to facilitate the learning process in target domains, achieving efficient adaptation and quicker mastery of new tasks. In the domain of reinforcement learning, Zhu et al. [23] identified that transferable knowledge can encompass demonstrated trajectories, model dynamics, teacher policies, and value functions. Furthermore, based on our literature review, the knowledge can be transferred across task domains, from simulation to real-world settings [10, 14], and from visual representations to policy formulations [13, 21, 24] within the reinforcement learning framework.

An exemplary method for transferring visual knowledge within the RL-based robotics areas is through contrastive unsupervised representation learning (CURL). This approach leveraged contrastive learning to derive visual knowledge and updated the visual encoder during the policy training process, effectively capturing and utilizing visual cues to inform policy decisions to improve improved generalization and sample efficiency [24]. In studies related to CausalWorld, Yoon et al. acquired structured and semantic visual representations from images via object-centric representation (OCR) pre-training, employing the encoded OCR to enhance reinforcement learning (RL) training [13]. Their evaluation focused on the efficiency of a single finger reaching a block. However, for more complex tasks, such as pushing and picking up a block with three fingers, the utility of the learned visual information proved limited. Our experiments with both CURL and OCR frameworks to extract visual information revealed that the agent was unable to learn, indicating that the visual information provided was insufficient for mapping to the optimal policy in tasks requiring intrinsic coordination and cooperation. For this study, state input containing the current and goal information for the task was used instead.

Our prior work in transfer reinforcement learning within the ship collision avoidance domain encompassed both feature and policy transfer [21, 22]. We developed an end-to-end framework that integrates a feature extractor with a policy network. The agent was pre-trained on relatively simple source tasks. Then, the whole or partially acquired knowledge was applied to target tasks of varying similarity and complexity. Despite incorporating rules, risk assessment, and dynamics in the environment, the input was solely visual, relying on 2D images, and the action space was coarsely discretized. This study builds upon our previous work by extending it to 3D scenarios, addressing more complex tasks such as robotic manipulation, and incorporating more realistic control mechanisms and dynamics.

### 2.4 Curriculum Learning

Curriculum learning is a strategic approach in machine learning that structures the learning process to mimic the way humans learn progressively. Several studies indicate that carefully designed curricula could lead to faster learning and the discovery of better policies compared to traditional RL approaches and enable learning in complex environments that were beyond the reach of standard RL methods [26, 27]. Narvekar et al. have identified key evaluation metrics for assessing the benefits of curriculum learning over non-curriculum RL approaches, including jump starts, time to threshold, and asymptotic performance [28].

Curriculum design is a critical challenge, and human participation plays a vital role in sequencing and evaluating intermediate tasks. The study most closely related to ours is the Guided Curriculum Learning by Tidd et al., which introduced a 3-stage curriculum for training robots to navigate complex terrains. The curriculum progressively increased in complexity while reducing guiding forces across the stages [31]. In our study, we increased the complexity but avoided using guidance as we observed the agent acquired skills in a manner distinctly different from our human in this task. Rather than pre-designing the curriculum, human feedback can be integrated into the learning process. Zeng et al. introduced a human-in-the-loop curriculum, developing an interactive platform that allows humans to offer curriculum feedback and adjust task difficulty levels, customizing the reinforcement learning process to achieve desired performance outcomes [32].

Beyond manual design, auxiliary networks are widely used to help generate curriculum automatically. A Teacher-Student Automatic Curriculum Learning framework was proposed by Campbell et al. [29]. The student network worked as an RL learner. The teacher network also worked following the Markov Decision Process and adjusted the curriculum based on the learning progress of the student network provided by gradient norms. Racaniere et al. introduced a judge model to predict the feasibility, besides the curriculum prediction model and the RL training model. The goals with different levels of validity and feasibility were sampled [30]. However, in our study, where goals are of similar difficulty, and the main challenge lies in the randomness introduced to the variables in each episode, these automatic methods may not be directly applicable. Nevertheless, they provide valuable insights for future research directions.

### 3. METHOD

In this section, we describe the methodology employed for our study. Our approach involves training an agent to obtain knowledge about the task using the Soft Actor-Critic (SAC) reinforcement learning algorithm [36]. To effectively transfer the acquired knowledge, we employed two distinct strategies: initially establishing a strong foundation by fine-tuning a pre-trained model, which is described in 3.2. Or addressing the hard task by constructing a curriculum that introduces complexity in increasing order, allowing the agent to build upon its skills incrementally through curriculum learning, as described in 3.3.

### 3.1 Training the Agent by Reinforcement Learning.

SAC, a well-known model-free RL method, was selected to train the agent for its efficiency and stability in solving complex decision-making problems. SAC is built on the actor-critic architecture, where the 'actor' is a policy network that outputs a probability distribution over actions, facilitating exploration and exploitation by sampling from this distribution. On the other hand, the 'critic' is a value network that evaluates these actions by estimating the Q-value. The clipped double-Q trick [33] is implemented, mirroring strategies in the Twin Delayed DDPG (TD3) [34], mitigating the overestimation issue of Q-value predictions commonly seen in deep RL.

Moreover, SAC integrates entropy regulation into the objective function and aims to optimize the cumulative rewards while also maximizing entropy. The objective function $J(\pi)$ of the policy $\pi$ is shown in Eqn. 1, where $\alpha$ is a coefficient of the trade-off between the reward $r(s_t, a_t)$ and the entropy $\mathcal{H}(\cdot | s_t)$, governing the randomness of the optimal policy. In contrast to traditional RL algorithms that aim to find a deterministic policy maximizing the expected return, SAC seeks a stochastic policy that balances the expected return and entropy. This results in a 'soft' policy update strategy, making the learning process more stable and less prone to getting stuck in local optima.

$$J(\pi) = \sum_{t=0}^{T} E_{(s_t, a_t) \sim \rho_\pi}[r(s_t, a_t) + \alpha \mathcal{H}(\cdot | s_t)] \quad (1)$$

The robot's actions are represented by a 9-dimensional array corresponding to the joint positions of each finger, sampled from a predefined continuous space. As for the state input, the environment provides visual inputs consisting of current and goal images captured from three viewpoints by cameras. Additionally, the structured observation is represented by a 56-dimensional array, which includes task-relevant variables such as the remaining time for the task, joint positions, joint velocities, and the linear velocities of blocks, among others, as illustrated in Figure 1. Initially, we attempted to utilize visual inputs, aiming to apply a methodology similar to that of our previous research [21, 22]. Our goal was to compare the results. However, despite experimenting with various CNN-based SAC models from Stable Baselines3 and other transformer-based visual encoders [13,24], and despite trying various modifications such as altering the number of input images, employing different preprocessing methods, varying the stacking techniques, and switching from images of different views to images from different time steps to create a new frame, all attempts were unsuccessful. This failure can be attributed to the difficulty in distinguishing the actions (i.e., the joint positions of the fingers) through images due to the complexity of the action space where numerous combinations can occur and the fact that the fingers are relatively small compared to the entire image. Consequently, we opted to leverage structured state input instead.

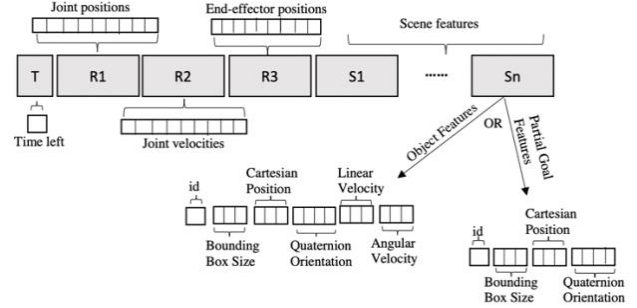

**FIGURE 1:** Illustration of structured state input [10]

A compound reward function from the original paper [10] was utilized, as shown in Eqn. 2. $A_{intersect}$ represented the intersection area between the current block position and the desired goal position, while $A_{union}$ represented the union of the two positions. $d(o, g)$ represents the distance between the block(object) and the goal. $d(o, e)$ represents the distance from the end-effector of the robotic arm to the block. The first component was designed to guide the agent in pushing the block toward the desired goal through sparse rewards. It became greater than 0 only when the block intersected with the goal, which was hard to achieve at the beginning. Meanwhile, the second component encouraged the agent to move the block towards the goal by reducing $d(o, g)$ at each time step, complementing the sparse reward with more continuous feedback. Similarly, the third component motivated the agent to move the finger closer to the block for grasping. The weights for each component were determined through reward engineering.

$$R = 100 * \frac{A_{intersect}}{A_{union}} - 250*(d^t(o,g) - d^{t-1}(o,g)) \\ - 750*(d^t(o,e) - d^{t-1}(o,e)) \quad (2)$$

The implementation was carried out using the PyTorch-based Stable Baselines3 library. The hyperparameter followed the original paper [10].

### 3.2 Fine-tuning

The case study task involves pushing a block to a specified goal, with the block's size, weight, or the goal's position being randomly sampled to increase the difficulty of the pushing task. The primary challenge in addressing this complex manipulation task stemmed from the agent's initial inability to push the block toward the goal. Because three fingers haven't mastered foundation skills such as grasping and moving the block with cooperation. Meanwhile, the sparse reward defined by the intersection portion dominated the total reward function. The agent will receive a significant reward only when the block intersects with the goal position. Thus, the experience buffer lacked experience with high reward initially, hindering the training process.

To address this issue, one effective strategy involves knowledge transfer, beginning with a pre-trained model as a foundation rather than starting from scratch. This approach, known as fine-tuning in the context of transfer learning, allows the agent to adapt more efficiently to the target task by leveraging prior knowledge. We tried two levels of knowledge transfer. The first one is the whole model transfer. The whole model was pre-trained in a simplified, straightforward pushing task, which took a very short time to achieve. The pre-trained parameter was utilized as the starting point for the target-pushing tasks with randomness embedment with the task variables. As illustrated in Figure 2, the actor network and critic network (including the state value network) were initialized with the pre-trained parameter and fine-tuned in the target tasks.

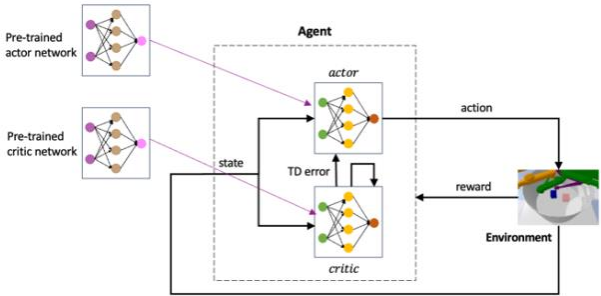

**FIGURE 2:** Whole model transfer and fine-tuning pipeline

The whole model transfer includes both the policy and value functions, which might be tightly coupled with the specifics of the source task environment, including its dynamics and reward structure. If the target task operates under significantly different dynamics or objectives, the value function from the whole model might not generalize well, potentially leading to suboptimal performance. Policy-only transfer, by focusing on the decision-making strategy independently of the value estimation, may offer better generalization across tasks with different dynamics. Also, it allows for the application of a learned strategy within a new reward context, offering the flexibility to adapt the policy to optimize for different reward signals without being constrained by the value function's assumptions about rewards. To achieve this flexibility, we employed policy transfer from the source task to the target task. As illustrated in Figure 3, the parameters of the actor (policy) network from the model pre-trained on the source task were transferred to the target network for fine-tuning on the target task while the critic network was randomly initialized. The outcomes of both approaches are discussed in Section 5.

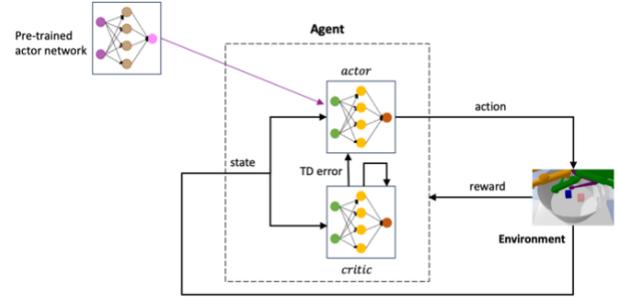

**FIGURE 3:** Policy transfer and fine-tuning pipeline

### 3.3 Designing the Learning Curriculum.
Although the fine-tuning method shows promise in enhancing learning efficiency, the necessity for pre-training should also be considered. To address this, we explored curriculum learning as a strategy to train the agent to tackle the challenging task in a single training session through designing an effective curriculum, eliminating the need for pre-training and fine-tuning.

As previously mentioned, in highly complex tasks, it is impossible for the agent to get close to the goal and receive the sparse reward in the early episodes. Our goal was to provide guidance that increased the likelihood of the agent obtaining this reward early, thereby facilitating the accumulation of valuable experience as swiftly as possible. Driven by this motivation, we broke down the original task into simpler sub-tasks of varying complexity to design the curriculum. The task that rapidly enhanced performance was identified through trials and implemented as the initial stage of the curriculum, termed the *preliminary task*. In this phase, all randomness was eliminated, and task variables were fixed to ensure a focused and effective learning environment. After acquiring knowledge in the first stage, randomness within a small range was introduced to the *intermediate task*. This step was designed to help the agent manage variations and achieve generalization. Subsequently, the level of randomness was gradually increased to the desired level in the *advanced task*, enabling the agent to fully master the original task. Throughout this process, knowledge was learned and transferred stage by stage.

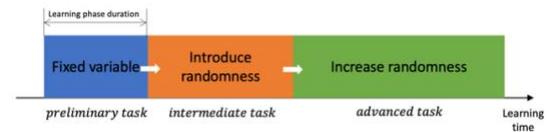

**FIGURE 4:** Illustration of curriculum design

The overall curriculum design is conceptually shown in Figure 4. Note that it doesn't exactly show the exact curriculum we used. In the experiments, the *intermediate task* may be skipped, the length of each stage may be varied as needed.

### 4. CASE STUDY DESIGN

The case study involved pushing a block to a designated goal, introducing randomness in both task-specific and environmental variables. During training, the agent was only provided with task-specific variables as state inputs for the neural networks, while changes in environmental variables, such as weight and friction, remained undisclosed. One might question the exclusion of environmental variables from the observation space. Our experimental findings indicated that incorporating environmental variables could detrimentally affect performance. This can be because, although environmental variables are relevant to the manipulation task, they are not as critical as task-specific variables that directly indicate the goal and current progress. Including environmental variables tended to distract the agent's focus from the primary task. Consequently, we designed tasks that were either context-aware by varying task-specific variables (such as size and goal positions) or context-unaware by altering environmental variables (like weight) to investigate both fine-tuning and curriculum learning methods. The challenge in the context-aware scenario was to adjust the policy to account for observable changes. In contrast, the context-unaware scenario demanded the development of a generalized policy capable of performing effectively across a range of unknown changes.

To ensure the task's difficulty level, we expanded the range of variables beyond the original task settings [10]. For $task$ 1, the block's length, width, and height were independently sampled from a range of 0.035 to 0.095 meters, potentially resulting in an irregular cuboid. $task$ 2 involved pushing the block to a goal position within a cylindrical coordinate system, with the position varying between (0, $-\pi$, 0.0375) and (0.15, $\pi$, 0.0375) meters and the orientation ranging from $-\pi$ to $\pi$ radians. $task$ 3 required pushing a block with a weight that varied from 0.015 to 0.5 kilograms. After trials and performance comparisons, the curriculum for each task was manually designed and is detailed in Table 1. The fixed value was applied to the $preliminary\ task$. For the $intermediate\ task$ and $advanced\ task$, the ranges provided were used to randomly sample the variables for each episode.

**Table 1. Curriculum details**

|  | $preliminary$ $task$ | $intermediate$ $task$ | $advanced$ $task$ |
|---|---|---|---|
| $task$ 1 (size) | 0.075 | [0.015, 0.095] | [0.015, 0.095] |
| $task$ 2 (goal) | (0.08, $\pi/2$, 0.0375) | [(0.04, 0, 0.0375), (0.12, $\pi$, 0.0375)] | [(0, $-\pi$, 0.0375), (0.15, $\pi$, 0.0375)] |
| $task$ 3 (weight) | 0.15 | [0.05, 0.25] | [0.015, 0.5] |

The timing of progressing through tasks is crucial. We aimed to investigate whether a sub-task should be fully mastered before advancing to the next level or if it is more effective to introduce the next level as soon as the agent shows some, but not perfect, progress on the current sub-tasks. We explored different timings to determine the most appropriate curriculum structure. Additionally, it's important to note that for the fine-tuning method, the model was pre-trained on the preliminary task to reduce pre-training time and then transferred to the original task (which is the same as the advanced task) for fine-tuning.

## 5. RESULTS AND DISCUSSION

### 5.1 Fine-tuning Results and Discussion

In this subsection, we present the results from fine-tuning case studies and discuss the features and implications of the results obtained from within-task and across-task cases.

#### 5.1.1 Fine-tuning within the tasks

Both whole network and policy transfer methods were implemented for context-aware scenarios in $task$ 1 (random size) and $task$ 2 (random goal), as well as for the context-unaware scenario in $task$ 3 (random weight). The model was pre-trained on pushing the block with fixed task parameters and subsequently fine-tuned for pushing blocks in a dynamic environment, incorporating variations in both task and environmental parameters, as described in Section 4. The cumulative reward obtained throughout the learning process is depicted in Figures 5 to 7. Each line in these figures was produced using 10 random seeds and has been smoothed for clarity.

We utilized the jump starts, time to threshold, and asymptotic performance evaluation metric [28] to access the learning process of the fine-tuning method against the baseline (learning from scratch). It was observed that whole network transfer altered the learning process in context-aware scenarios for $task$ 1 and $task$ 2 by providing jump-starts and reducing the time to convergence, significantly enhancing learning efficiency. Furthermore, the asymptotic performance of the whole model transfer method exceeded that of the baseline in $task$ 2, demonstrating that the whole model transfer and fine-tuning improved the learning process by knowledge reuse and adaptation. However, the policy-only transfer method did not show significant improvement, indicating that the policy network alone lacked sufficient knowledge to boost the learning process. This may also suggest a tight coupling between the policy network and the value network during the update process; thus, transferring only the policy without value evaluation does not accelerate convergence.

In $task$ 3, which is a context-unaware scenario, different from the context-aware scenario, the policy-only transfer demonstrated better performance than the whole model transfer and slightly better than the baseline. The potential reason is that, despite changes in weight in the target task, these were not observed by the agent due to the unawareness setting. The agent initially treated it similarly to the source task, and the pre-learned policy provided some initial insight. However, while the whole network transfer achieved jump starts in the early stage, it experienced significant oscillations and performed much worse compared to the baseline. This may be due to the fact that the difference between the source task and the target task (weight variation) was not observable by the model, leading to inaccurate value estimations based on the observations. Moreover, the pre-trained whole network, having converged to the optimal solution in the source task, may have caused the fine-tuning process to become stuck in local optima.

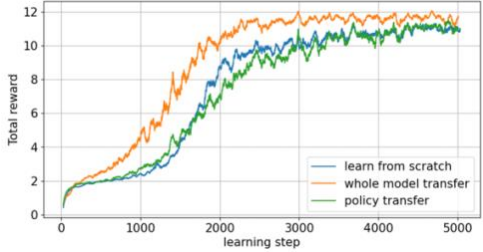

**FIGURE 5:** $task$ 1 fine-tuning learning process

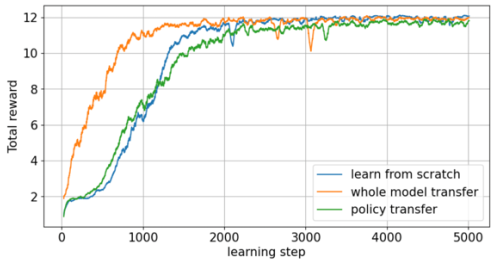

**FIGUR 6:** $task$ 2 fine-tuning learning process

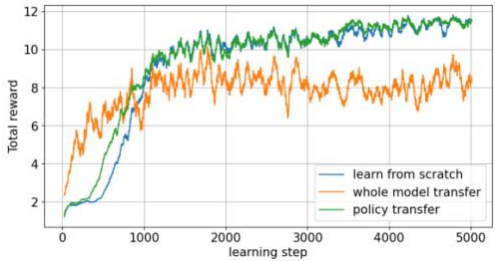

**FIGURE 7:** $task$ 3 fine-tuning learning process

### 5.1.2 Fine-tuning across the tasks

As humans, we are often able to easily learn a complicated task based on a related but simpler task. For example, we can master pushing blocks to arbitrary goals quickly as an add-on skill if we already know how to grasp a block. Inspired by this observation, we sought to determine whether robot learning could follow a similar logic. If so, this could lead to a more efficient training pipeline. Additionally, pre-training the robot to grasp a block required significantly less time than pre-training it to push a block to a fixed goal in the fine-tuning experiment above.

The pre-training task was designed to have the robot grasp the block from the same initial position as in the pushing task. The desired outcome was for the distance between the three fingers and the block to be zero, with minimal movement of the block, ensuring it does not move to an unintended location before the goal is known. The reward function, defined in Equation 3, uses $\|o^t - o^{t-1}\|$ to represent the distance between the current and previous block locations, serving to limit block movement. This reward function design closely mirrored that of the pushing task in Eqn. 2, aiming to maintain a similar reward function structure and value estimation distribution.

$$R = -100 * d^t(o,e) - 250 * \|o^t - o^{t-1}\| \\ - 750 * (d^t(o,e) - d^{t-1}(o,e)) \quad (3)$$

The results, presented in Figure 8, are labeled as "across-task" to indicate that the agent was pre-trained on grasping and then fine-tuned for pushing. These results were compared with the previous fine-tuning outcomes shown in Figure 6, which are labeled as "within-task" transfer in this comparison. We observed that across-task fine-tuning achieved performance very similar to that of within-task fine-tuning but required only one-tenth of the pre-training time. From this, we can deduce that, from the knowledge acquired during the fixed-goal-pushing pre-training, only the grasping skill was applicable to the random-goal-pushing task. The skill of moving with forward momentum could be developed based on the learned grasping skill during the fine-tuning process. This suggests that the entire pushing task can be broken down into simpler pre-training and fine-tuning stages. Furthermore, this finding provides us with future research directions, as other complex tasks like picking, placing, and stacking blocks may also benefit from following this training pipeline.

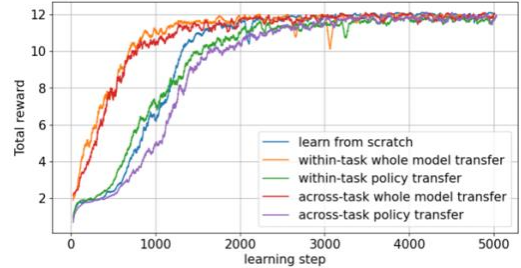

**FIGURE 8:** $task$ 2 3-stage-curriculum learning process.

### 5.2 Curriculum Learning Results and Discussion.

3-stage curriculum learning following the curriculum in section 4 has been applied to $task$ 1, as shown in Figure 9. By observing the learning process, we can see that learning efficiency was boosted a lot in the earlier stage compared to the baseline. This improvement is attributed to the agent focusing on simpler sub-tasks, such as pushing the block with a fixed size or minor shape variations, which resulted in more valuable experiences being stored in the experiment buffer. However, performance experienced a drop each time the sub-task was changed, indicating that the timing of sub-task transitions significantly impacted learning outcomes. Furthermore, while 3-stage learning enhanced learning effectiveness in the initial episodes, it did not lead to improvements in the time to threshold or asymptotic performance due to two instances of performance decline.

 

To achieve faster convergence, a 2-stage curriculum learning (skipping the *intermediate task*) strategy was also applied to *task* 1. This approach was chosen because it resulted in only one performance drop during the learning process. The timing of the transition to the next sub-task became even more critical, as depicted in Figure 10. Although the performance drop in the 2-stage curriculum was more pronounced than in the 3-stage curriculum—due to a larger increase in complexity—the optimal timing for changing sub-tasks (at 1200 timesteps) significantly reduced the time to converge and enhanced the asymptotic performance. Changing the sub-task at 800 timesteps proved too early, as insufficient knowledge had been acquired and performance had not reached a sufficiently high level. Conversely, transitioning at 1800 timesteps was too late, as prolonged engagement with a simpler task negatively impacted the agent's ability to achieve as high a final reward when faced with a more complex task compared to the optimal timing.

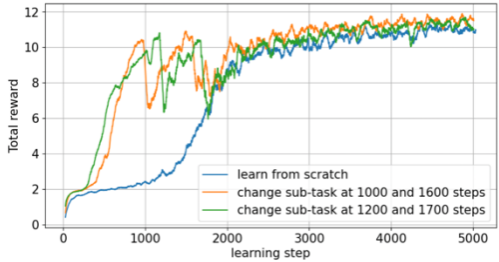

**FIGURE 9:** *task* 1 3-stage-curriculum learning process.

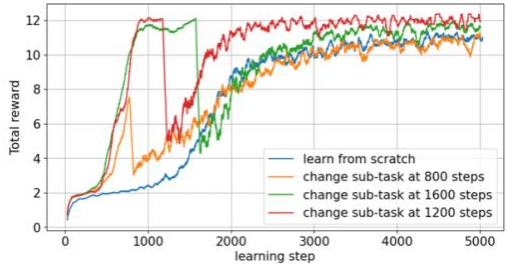

**FIGURE 10:** *task* 1 2-stage-curriculum learning process.

The top performers from both the 3-stage and 2-stage curriculum learning approaches were compared with the best candidate from the fine-tuning method. The results are presented in Figure 11. All learning approaches surpassed the baseline, highlighting the advantages of knowledge transfer. Specifically, the 2-stage curriculum, with the sub-task transition occurring at 1200 timesteps, emerged as the optimal solution for pushing blocks of varied sizes.

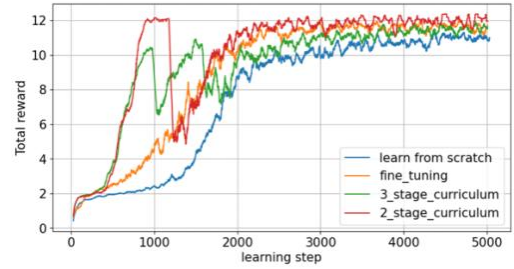

**FIGURE 11:** Comparing the best options of each learning strategy in *task* 1

We applied a similar procedure to *task* 2 and *task* 3 to identify the optimal curriculum design and sub-task transition timing and to compare these with the fine-tuning method in order to determine the most suitable overall learning strategy for each task.

Contrary to *task* 1, the results for *task* 2, as shown in Figure 12, indicate that the 3-stage curriculum outperformed the 2-stage curriculum. The gradual increase in complexity led to improved learning performance with respect to goal position randomization. However, none of the curriculum learning candidates showed improvement in terms of total reward. The optimal solution for pushing a block to an arbitrary goal was found to be fine-tuning the pre-trained model. Although the curriculum led to an increase in rewards when reaching the fixed goal in the $p$, this effect disappeared when dealing with random goals in more challenging curriculums. This suggests that changing the goal position significantly alters the task process, requiring the agent to spend more time fine-tuning rather than quickly mastering it within a curriculum with limited learning time.

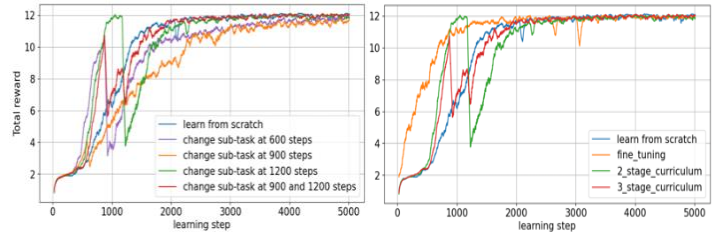

(a) 2-stage-curriculum learning process.  (b) Comparison of the best option Of each learning strategy

**FIGURE 12:** Curriculum learning and comparisons in *task* 2

As the result for context-unaware *task* 3 shown in Figure 13, the best solution is 2-stage curriculum learning. As we discussed in 5.1.1, the fine-tuning method didn't improve the learning process well due to the difference between the source task and the target task being unobserved. For curriculum learning, unlike in *task* 1, introducing the advanced task BEFORE the agent fully mastered the preliminary task yielded the best results in terms of convergence speed and the final reward as shown with the red line. Since the change was unobserved, performance did not suffer the drop typically seen in context-aware scenarios. Introducing an *intermediate task* midway through the process (indicated by the purple line) did not enhance efficiency compared to the early-stage progression of the 2-stage curriculum (indicated by the red line). This is because adding the *intermediate task* at 400 steps in the 3-stage curriculum had a similar effect to introducing the advanced task at the same point in the 2-stage curriculum. Furthermore, the performance drop associated with progressing to the *advanced task* was not alleviated by the inclusion of the *intermediate task* in the 3-stage curriculum. However, it did help mitigate the performance drop compared to when the *advanced task* was introduced directly at 800 steps (indicated by the orange line).

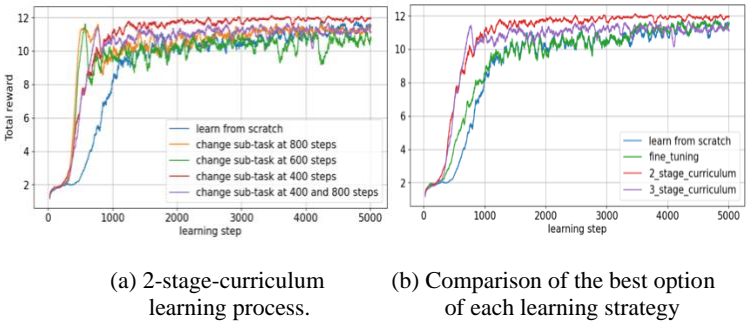

(a) 2-stage-curriculum learning process.

(b) Comparison of the best option of each learning strategy

**FIGURE 13:** Curriculum learning and comparisons in *task* 3

### 5.3 In-distribution and Out-of-distribution Evaluation

The best option from each learning method was evaluated over 100 episodes for both in-distribution and out-of-distribution tests. The out-of-distribution tests extended the randomness range of *task* 1, *task* 2 and *task* 3 by approximately 67%, 13%, and 21%, respectively, representing the maximum applicable value for each task. The mean and standard deviation of the fractional success for each task are presented in Figure 13.

From the results, we can find that except for *task* 2, where the curriculum learning method was not applicable, all the transfer learning methods achieved equivalent generalization capability or surpassed the baseline. It demonstrated that the knowledge can be transferred through pre-training and fine-tuning or within the curriculum.

The evaluation rankings were highly consistent with the order of rewards, indicating a strong correlation between learning efficiency and generalization. For example, *task* 1, the best solution 2-stage curriculum learning selected in 5.2 has also been proved to be the most robust and generalizable.

There were some inconsistencies between training rewards and testing results, highlighting the impact of transfer learning from various perspectives. In *task* 2, although 3-stage learning did not show a clear efficiency improvement, as depicted in Figure 11, the evaluation results demonstrated enhanced generalization, indicating that the 3-stage curriculum contributed to better generalization. Also, it was surprising to observe that although the best fine-tune candidate (policy-only transfer) of *task* 3 didn't show much improvement regarding the learning process, it exhibited better generalization compared to the baseline and 3-stage curriculum for both in-distribution and out-of-distribution tests.

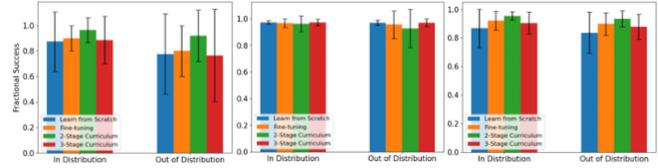

**FIGURE 13:** Mean and Standard Deviation of Fractional Success: In-Distribution and Out-of-Distribution.

### 6. CONCLUSIONS AND FUTURE WORK

This study presented a comprehensive examination of the use of knowledge transfer strategies, namely fine-tuning and curriculum learning, in the context of robotic manipulation tasks utilizing a tri-finger robotic arm. Through extensive experiments within the CausalWorld platform, we demonstrated the effectiveness of these strategies in enhancing learning efficiency and robustness in complex manipulation tasks. The key findings of our work are summarized as follows:

- *Whole network transfer vs. policy-only transfer:* The whole network transfer demonstrated superiority over policy-only transfer in scenarios requiring context-awareness. This highlights the interdependence between the policy network and the value network throughout the learning process.

- *Cross-task fine-tuning efficiency*: Cross-task fine-tuning proved to be highly efficient, achieving comparable training outcomes to within-task fine-tuning while significantly reducing the need for pre-training time.

- *Context-unaware scenario adaptability*: In context-unaware scenarios, transferring the complete model from a source task proved detrimental to learning, likely due to unobserved environmental variations. However, policy-only transfer was not only more effective but also exhibited enhanced generalization capabilities.

- *Curriculum learning design parameters*: In curriculum learning, the sequencing of learning stages, the relevance of sub-task contexts, and the timing of transitions emerged as pivotal factors for successful task mastery.

- *Sub-task transition timing*: Premature transitions to subsequent sub-tasks hindered the agent's knowledge acquisition. Conversely, excessive delays caused an overfitting to simpler tasks, underscoring the importance of well-timed progressions.

- *Stage-based learning efficacy*: A 2-stage learning approach accelerated the learning process, allowing the agent to apply basic skills to more complex real tasks promptly. In contrast, a 3-stage curriculum ensured a more gradual and smooth learning experience, ultimately leading to better generalization.

These findings showed the key factors of each learning method and their effects. They also highlight the strategy of selecting the appropriate learning based on various task conditions, providing some insights for designing training procedures for complex manipulation tasks. Building on the insights gained from this study, our future research endeavors will focus on several directions:

- *Extended application of cross-task fine-tuning*: Given the promising results observed with cross-task fine-tuning, we plan to explore its application to a broader range of complex robotic manipulation tasks. Investigating the transferability of skills across different task domains could further enhance learning efficiency and generalization.

- *Automatic curriculum generation*: While manual curriculum design has proven effective, it requires significant effort and domain knowledge. Future work will explore automated methods for curriculum generation, aiming to dynamically adjust the learning process based on the agent's performance. This could lead to more adaptable and scalable learning strategies for complex engineering problems.

- *Broader exploration of knowledge transfer and curriculum learning*: We intend to apply the principles of knowledge transfer and curriculum learning to other engineering scenarios beyond robotic manipulation. By exploring these strategies in diverse applications, we aim to uncover universal guidelines that can facilitate the development of advanced learning-based solutions.

**ACKNOWLEDGEMENTS**

We adapted the CausalWorld code by extending the intervention space, incorporating customized tasks, interventions, curriculums and evaluation pipeline, and modifying the reinforcement learning algorithm to suit our case study. The code and implementations can be provided by the corresponding author upon reasonable request.

This paper is based on the work supported in part by the Autonomous Ship Consortium (ASC) with members of BEMAC Corporation, ClassNK, MTI Co. Ltd., Nihon Shipyard Co. (NSY), Tokyo KEIKI Inc., and National Maritime Research Institute of Japan. The authors are grateful for their support and collaboration on this research.